\documentclass{article}

 \usepackage[preprint]{neurips_2026}

\usepackage[utf8]{inputenc} 
\usepackage[T1]{fontenc}    
\usepackage{hyperref}       
\usepackage{url}            
\usepackage{booktabs}       
\usepackage{amsfonts}       
\usepackage{nicefrac}       
\usepackage{microtype}      
\usepackage{xcolor}         
\usepackage{amsmath}
\usepackage{dsfont}
\usepackage{enumitem}
\usepackage{wrapfig}
\usepackage{comment}
\usepackage{graphicx}
\usepackage{color}
\usepackage{subcaption} 
\usepackage{geometry}

\usepackage{booktabs}
\usepackage{multirow}

\usepackage{subcaption}
\usepackage{wrapfig}

\title{Neural-Behavioral Representation \\ of Natural Whole-body Movement in Monkeys}

\author{
  Jieshi He$^{1}$\thanks{Equal contribution}
  \And
  Puzhe Li$^{1}$\footnotemark[1]
  \And
  Yanan Sui$^{2}$\thanks{Corresponding to: Yanan Sui <\texttt{ysui@tsinghua.edu.cn}> and Mu-ming Poo <\texttt{mpoo@ion.ac.cn}>}
  \And
  Mu-ming Poo$^{1}$\footnotemark[2] 
  \AND
  \normalfont $^1$Center for Excellence in Brain Science and Intelligence Technology, CAS \\ $^2$Tsinghua University
}

\begin{document}

\maketitle

\begin{abstract}

Understanding how cortical activity represents natural whole-body behaviors in primates remains challenging. Limited by the diversity of movements and inaccessibility of large-scale neural representation of whole-body kinematics, previous motor decoding studies focused on constrained tasks and limited limb movements. Here, we present a neural-behavioral recording and modeling framework for freely moving monkeys, combining large-scale epidural cortical signals from distributed sensory- and motor-related areas with synchronized multi-view motion capture through a custom-made data collection platform. We reconstructed whole-body monkey kinematics and learned a compact behavior prior using an autoregressive encoder-decoder model. Conditioned on neural signals, the model decoded accurate and realistic whole-body movement without explicit physical constraints. Our results provide a novel proof-of-concept approach for decoding natural whole-body movements in primates using large-scale intracranial neural activity.

\end{abstract}

\section{Introduction}
\label{sec:intro}

The cerebral cortex supports sensory processing, motor planning, movement execution, and adaptive behavioral control. A central goal in computational and systems neuroscience is to understand how cortical population activity represents movements and to use the representations to reconstruct behaviors from neural activities  \cite{borgognon2025regional, umeda_decoding_2019, moritz_direct_2008, laurence2023robust}. Substantial progress has been made in decoding intended or executed movements from cortical activity, particularly for upper-limb reaching, hand grasping, and lower-limb locomotion under constrained experimental paradigms \cite{ottenhoff2025decoding, haghi_enhanced_2024}. However, most existing studies decode restrained postures or limited body segments, relying on predefined tasks and repeated trials. The neural representations underlying natural whole-body movements remain largely unexplored. A few studies have used neural features extracted from specific brain regions to classify constrained whole-body movement categories in monkeys, such as walking, jumping, or resting \cite{testard2024neural, voloh2023hierarchical}. Yet no prior report has shown direct reconstruction of continuous whole-body movements of primates from cortical neural representations.

Natural whole-body movement poses challenges that are qualitatively different from those in conventional motor decoding experiments \cite{LEOPOLD2024102913}. First, large-scale body movement introduces substantial artifacts into neural recordings \cite{MORAN2010741}, making it difficult to simultaneously acquire reliable brain signals and natural movement data. Second, compared with constrained settings, freely moving monkeys exhibit much richer behavioral patterns  \cite{lanzarini2025neuroethology}. Reconstructing whole-body movement requires predicting kinematics of all body parts, leading to a substantially higher-dimensional output space. Third, natural movements are spontaneous, irregular, and weakly repeated. Unlike instructed laboratory tasks, they do not provide aligned trials that can be averaged to improve signal-to-noise ratio. In constrained experimental paradigms, movements are usually triggered by task instructions, and are repeated in regular patterns. Moreover, similar movements may correspond to different neural activities under different contexts, while free movement also involves abundant spontaneous neural activity unrelated to the movement \cite{lanzarini2025neuroethology, weinreb2026spontaneous}. These factors make it difficult to learn stable neural-behavioral mappings using standard discriminative decoding approaches.

\begin{figure}
  \centering
  \includegraphics[width=\textwidth]{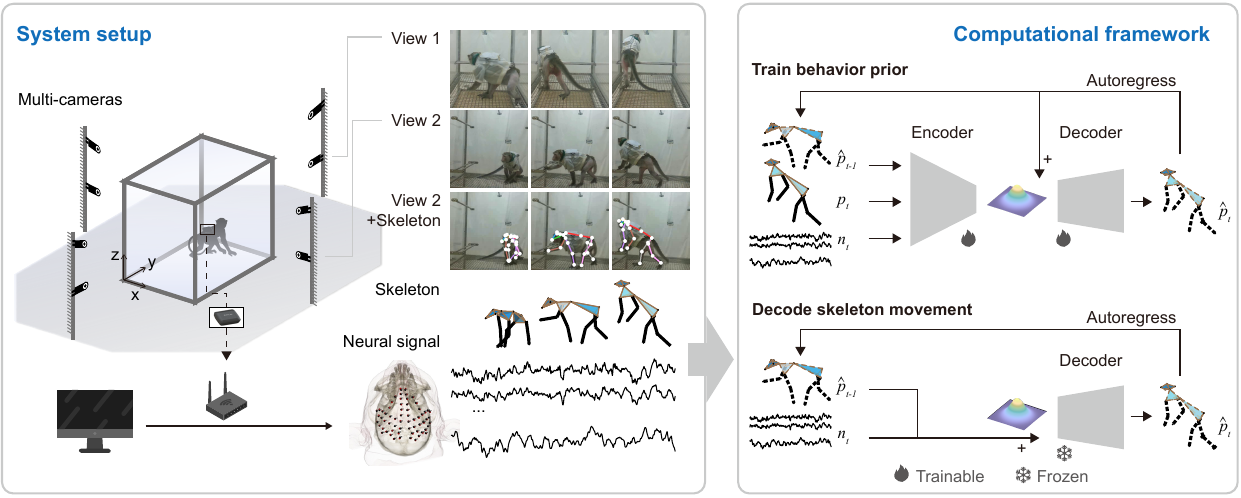}
  \caption{Pipeline for decoding natural whole-body movement from epidural neural signals. \textbf{System setup}: Whole-body movement data were captured using an 8-camera system and reconstructed into 3D skeleton. Synchronous recording of epidural neural signals from 62 channels were recorded and transmitted wirelessly. \textbf{Computational framework}: Collected data were divided into training and test sets. First, we trained a behavior prior on the training set by compressing movement data and neural signals into a latent space. Latent variables were then sampled and combined with the predicted state $\hat{p}_{t-1}$ and the neural signals ${n}_{t}$, and further passed through the decoder to reconstruct the current state $\hat{p}_{t}$. During testing, only the decoder was used, and the movement states were reconstructed in the same autoregressive manner as during the training.}
  \label{Fig:pipeline}
  \vspace{-12pt}
\end{figure}

In this work, we have addressed these challenges by developing a neural-behavioral recording and modeling framework for freely moving monkeys as shown in Fig. \ref{Fig:pipeline}. We implanted epidural electrodes over distributed motor-related cortical areas, enabling recording of cortical population activity. Compared with intracortical microelectrode arrays, these epidural recordings sacrificed single-neuron resolution for improved long-term stability and reduced tissue damage \cite{mueller2025mechanically,branco2023nine}. Compared with conventional ECoG arrays, our skull-fixed design enables broader cortical coverage while preserving the natural buffering structure of the skull, dura, and brain, which is important for reducing movement-induced artifacts during free-moving behaviors \cite{MORAN2010741}. 

To capture movement behavior, we build a synchronized multi-view motion capture system with eight cameras and reconstruct three-dimensional monkey kinematics during spontaneous movement. We defined a monkey kinematic model and converted tracked 3D joint trajectories into a compact whole-body movement representation. On top of this recording system, we designed the Neural-Behavioral Model, which learns a compact behavior prior from whole-body kinematic sequences while simultaneously learning how epidural neural activity represents the natural movement. During inference, the model reconstructs whole-body movement from the previous predicted state and the current neural signals. This formulation is designed for natural behavior, where movements are constrained both by biomechanical regularities and by time-varying prior cortical activity. Unlike direct regression models that map neural features to joint coordinates independently, our approach explicitly integrates neural representations with a learned behavior prior, enabling more realistic and temporally coherent movement prediction without imposing explicit physical constraints.

Experiments on synchronized neural and behavioral recordings from freely moving monkeys demonstrate the feasibility of reconstructing continuous whole-body movement from distributed epidural cortical activity. The Neural-Behavioral Model reconstructs realistic whole-body movement over multi-second time scales and outperforms both a behavior-only generative model and a LSTM-based model \cite{eyvazpour2026general}. These results show that neural signals provide behavior-specific information beyond generic behavior priors, which are essential for maintaining physically plausible whole-body posture and coordination in the high-dimensional movement space.

\textbf{Contributions.} \\
1) We have developed a neural-behavioral recording platform that synchronously captures large-scale epidural cortical activity and multi-view 3D whole-body kinematics in freely moving monkeys.\\
2) We have formulated natural whole-body movement reconstruction as a problem of neural-behavioral-conditioned autoregressive generative modeling, providing an ML framework that integrates distributed cortical representations with learned behavior priors.\\
3) We have provided the first demonstration, to our knowledge, that continuous and realistic whole-body movement of freely moving primates can be predicted from cortical activity under naturalistic conditions.

\section{Related work}
\label{section2}

\subsection{Predicting movement from neural signals}
Previous studies have shown that neural signals from the human primary motor cortex can decode movement intention or limited actual motor behaviors, such as arm reaching and hand grasping \cite{jang_decoding_2022, haghi_enhanced_2024, hotson_individual_2016, benabid_exoskeleton_2019, sliwowski_decoding_2022,wang_eeg-based_2022, thomas_decoding_2019}.  Similarly, neural recordings from the motor cortex of non-human primates have enabled prediction of upper-limb movements \cite{wessberg2000real,schaeffer_switching_2016, shimoda_decoding_2012, chao2010long, colins2024motor}. In such studies, subjects were required to perform predefined tasks repeatedly in a structured manner, which reduced noise and facilitated the discovery of stable neural-behavioral mappings. Most previous studies have adopted such approaches for supervised discriminative models \cite{eyvazpour2026general, farrokhi_state-based_2020}. Some works \cite{ye2021ndt, ye2023neural, rouanne_unsupervised_2022} have also explored unsupervised pretraining on large-scale neural data for downstream tasks, but these efforts were largely limited to simple classification or single-limb regression tasks.

\subsection{Neural-behavioral representation}
Some studies on neuronal spike data have explored cross-modal and cross-subject neural representations. The NDT2 \cite{ye2023neural} introduces Transformer-based pretraining for neural spike modeling by integrating data across subjects, tasks, and recording sessions to learn general-purpose neural representations. The results showed that large-scale multi-context pretraining substantially improved downstream decoding performance and adaptation to unseen scenarios.  A framework \cite{zhu2025neural} established reliable cross-modal representational alignment, enriching neural coding insights and enabling zero-shot behavior decoding. A study has demonstrated that applying orthogonal regularization to the latent movement space can significantly improve decoding accuracy across sessions, subjects, and paradigm generalization tasks \cite{tian2026multi}.

\subsection{Neural and behavioral dataset and setup}
The UniMo4D \cite{wang2026x} dataset provides large-scale cross-species 3D movement data but does not include neural activity. Previously reported neural-behavior recording systems include those that only capture whole-body movement without neural recording such as BigMaQ \cite{martini2026bigmaq}, or setups that recorded distributed neural signals together with motion capture of body parts and those recorded local neural signals with whole-body movement (see Table \ref{Table: dataset} in the Appendix). OpenMonkeyStudio \cite{bala2020automated} reconstructed 3D kinematics of 13 joints in moving monkeys while recording tens to hundreds of neurons per session. A study \cite{borgognon2025regional} performed microelectrode array (MEA) recordings from three brain regions during locomotion tasks, synchronized with EMG and limited limb kinematics. NeuroTycho \cite{bala2020automated} provided epidural ECoG signals with upper-limb kinematics under restrained conditions. In contrast, our work uniquely combined population-level neural recordings over multiple sensory and motor cortical areas with synchronously monitored whole-body 3D kinematics in freely moving monkeys.

\section{System setup}
 \label{section3}

To reconstruct natural whole-body movement from distributed cortical activity, we first need to collect freely moving behavior data together with synchronized large-scale neural signals. In this section, we describe our experimental setup, including epidural recording of neural signal, video acquisition of monkey behaviors, reconstruction of movement trajectories from video recordings, and the kinematics-representing model of whole-body movement.

\subsection {Data collection} 

\paragraph{Animal care and surgery} One adult female monkey (\emph{Macaca fascicularis}, 14 years of age) was used in this study. All experimental procedures were approved by the Ethics Committee of the relevant institutional review board (Protocol No.: NZXSP-2022-11-1). The animal was well cared in a certified experimental facility for macaque monkeys. Following a pre-surgical health assessment, the animal underwent implantation surgery of epidural electrodes. Sixty-four implantation sites were designated on the skull across bilateral sensory- and motor-related cortical areas based on pre-surgical CT and MRI scans, including the primary motor cortex (M1), primary somatosensory cortex (S1), premotor cortex (PMC) and superior and inferior parietal lobes (SIPL), two of which served as reference and ground electrodes. The electrodes, which are located at the tip of insulated metal screws with a length slightly exceeding the local skull thickness, contacted the dura surface to record intracranial population neuronal activity. These recorded signals exhibited properties comparable to those obtained by ECoG, while providing additional advantages over ECoG, including reduced invasiveness, greater flexibility in selecting recording regions, and broader cortical coverage. 

\begin{wrapfigure}{r}{0.5\textwidth}
    \centering
    \includegraphics[width=0.5\textwidth]{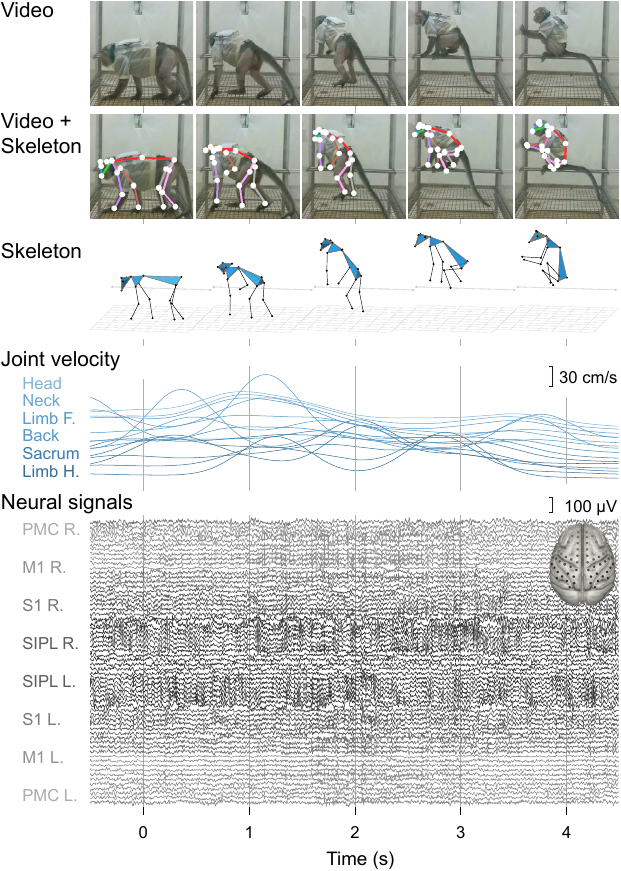}
    \vspace{-11.5pt} 
    \caption{Example captured movements and synchronized neural signals lasting 5 seconds. From top to bottom: raw video frames; video frames overlaid with tracked reconstructed 3D skeletons; 3D visualization of the skeleton; velocities of joints (blue color scale from light to dark indicates body positions from anterior to posterior; Limb F., forelimb; Limb H., hindlimb); and preprocessed neural signals from 62 channels. Gray color scales denote different cortical regions (R., right hemisphere; L., left hemisphere). The cortical locations of the corresponding channels are indicated at the corner. }
    \vspace{-20pt}
\label{Fig:data}
\end{wrapfigure}

\paragraph{Acquisition of neural signals and behavioral data} The recording session began three weeks after surgical implantation of electrodes. The animal was placed in a transparent observation cage (70 cm × 100 cm × 100 cm) in a dedicated behavioral recording room (3.5 m × 2.5 m) for multi-view behavioral video recording. The cage was equipped with horizontal bars at both upper and lower levels to support sitting, climbing, and swinging movements, and 8 cameras (Intel RealSense D435i) mounted on four walls of the room and provided comprehensive multi-view coverage of monkey movements. A wearable brain signal recording device (Quanlan RSC64) was placed in the backpack of the monkey jacket and connected to the lead wires. The animal was placed in the cage and allowed to move freely throughout sessions. Each behavioral experiment day included three sessions, each session lasting 30 minutes.

All 8 cameras were triggered simultaneously via a custom-designed program, with the first video frame of every camera and the onset of neural recording precisely aligned in time. Data were acquired continuously at stable sampling rates (videos: 30 Hz, neural signals: 2000 Hz). Rich behavioral repertoires of the monkeys were recorded in the experimental cage, including walking, climbing, swinging, scratching, and other spontaneous movements. We collected approximately 4 hours of synchronized neural and movement data from the monkey in 3 days.

The Fig. \ref{Fig:data} illustrates a monkey walking and stepping onto a bar. The figure shows raw video, reconstructed skeletons, joint velocities, and preprocessed neural signals. During quadrupedal locomotion, the left and right limbs exhibited alternating movement patterns, as reflected in velocity fluctuations. Synchronously, contralateral cortical hemispheres showed alternating high- and low-frequency oscillations. Consistent with prior studies \cite{Ray11526, Miller2424}, high-frequency activity in motor-related cortical areas was associated with movement execution. Time–frequency analysis of signals from primary motor cortex further supports this observation. Unlike findings under constrained conditions, this association was not consistently observed, as neural recordings during free movement contained more abundant spontaneous activity unrelated to movement (see Fig. \ref{Fig:PSD} in the Appendix for details).

\subsection{Kinematic representation of monkey movement}

\paragraph{Joint definitions} We defined 20 body keypoints of the monkey, including head, nose, left ear, right ear, neck, left shoulder, left elbow, left wrist, right shoulder, right elbow, right wrist, back, sacrum, left thigh, left knee, left ankle, right thigh, right knee, right ankle, and tail tip. We tracked and reconstructed the 3D coordinates of all keypoints in the world coordinate frame. Among these, we define 15 keypoints (nose, neck, left shoulder, left elbow, left wrist, right shoulder, right elbow, right wrist, sacrum, left thigh, left knee, left ankle, right thigh, right knee, and right ankle) including nose as "joints" for the kinematic representation, although the nose is not anatomically a true joint.

\paragraph{2D tracking of keypoints and 3D reconstruction of their coordinates} We first track keypoints in video recordings from the view of the 8 cameras. We used a clustering-based strategy to select 200 representative video frames from each camera for manual annotation of keypoint coordinates. A "top-down" pose estimation pipeline from DEEPLABCUT \cite{mathis2018deeplabcut} was adopted to identify keypoints in the video, which helped avoid spurious detection outside the cage caused by reflections from the acrylic cage walls. Specifically, we first performed monkey-level detection to localize the animal, and then estimated keypoint coordinates within the detected region. Camera calibration was performed using a standard checkerboard, and the resulting calibration parameters were used to reconstruct 3D joint coordinates in the world coordinate system (defined as shown in Fig. \ref{Fig:pipeline}) with the Anipose toolbox \cite{karashchuk2021anipose}. We imposed fixed bone-length constraints for all limb segments except the Neck-Back and Back-Sacrum connections, since the Back point is located on the animal’s jacket and thus may shift relative to the underlying body. During triangulation, we further applied temporal smoothing to joint trajectories to reduce tracking-induced jitter and non-physical discontinuities.

\paragraph{Kinematic representation of movement} We first defined a root coordinate system. The origin of root frame was set at the Sacrum joint in the world coordinate system, and the direction from Sacrum to Neck defined the z-axis. The projection of the z-axis onto the world x-y plane was then rotated counterclockwise by $90^\circ$ to define the y-axis, while the x-axis was determined by the right-hand rule. Since root rotation was typically the most difficult component to decode in movement prediction tasks, this definition constrained the rotational degrees of freedom of the root frame to two, thereby mitigating the decoding difficulty.

We then computed analytical solutions for joint rotations associated with joint trajectories using inverse kinematics, in contrast to numerical optimization methods employed in human models such as \cite{SMPL:2015}. We assumed fixed limb lengths during modeling and a constant limb length for each segment was estimated by averaging the lengths across the entire recording session. In addition, each child joint was constrained to two rotational degrees of freedom by removing rotation around the limb axis. The joint rotations were represented by Euler angles rather than rotation matrices. To avoid discontinuities caused by angle wrapping, we used the sine and cosine values of the Euler angles as the final rotation representation.

Next, We represent the state of a moving monkey as a matrix $P$, defined as

\begin{equation}
  P = (t_{root},r_{root},j^p,j^r,j^v)
  \label{define}
\end{equation}

Which consists of the displacement ($t \in \mathbb{R}^3$) of the root frame relative to the previous time step, the rotation ($r \in \mathbb{R}^2$) of the root frame relative to the previous time step, joint positions ($j^p \in  \mathbb{R}^{3*15}$) in the current root coordinate frame, joint rotations ($j^r \in \mathbb{R}^{4*14}$) in the current local joint coordinate frames, and joint velocities ($j^v \in \mathbb{R}^{3*15}$) expressed in the previous root coordinate frame. This yielded a unified state representation $P \in \mathbb{R}^{152}$. 

The sampling rate of reconstructed movement data from original video was 30 Hz. To increase the amount of novel neural information associated with each prediction step, we downsampled the data from 30 to 10 Hz, thereby reducing redundancy between consecutive neural signal windows. We used the predicted $t_{root},r_{root},j^r$ , together with the root-frame coordinates and rotation matrix from the previous time step, to compute the current joint coordinates via forward kinematics.

\section{Experiment}

In this section, we trained a behavior prior using a condition-guided autoregressive encoder-decoder model (Neural-Behavioral Model), and reconstructed monkey movements from neural signals. We then conducted a systematic evaluation of the reconstruction performance.

\begin{figure}
  \centering
  \includegraphics[width=\textwidth]{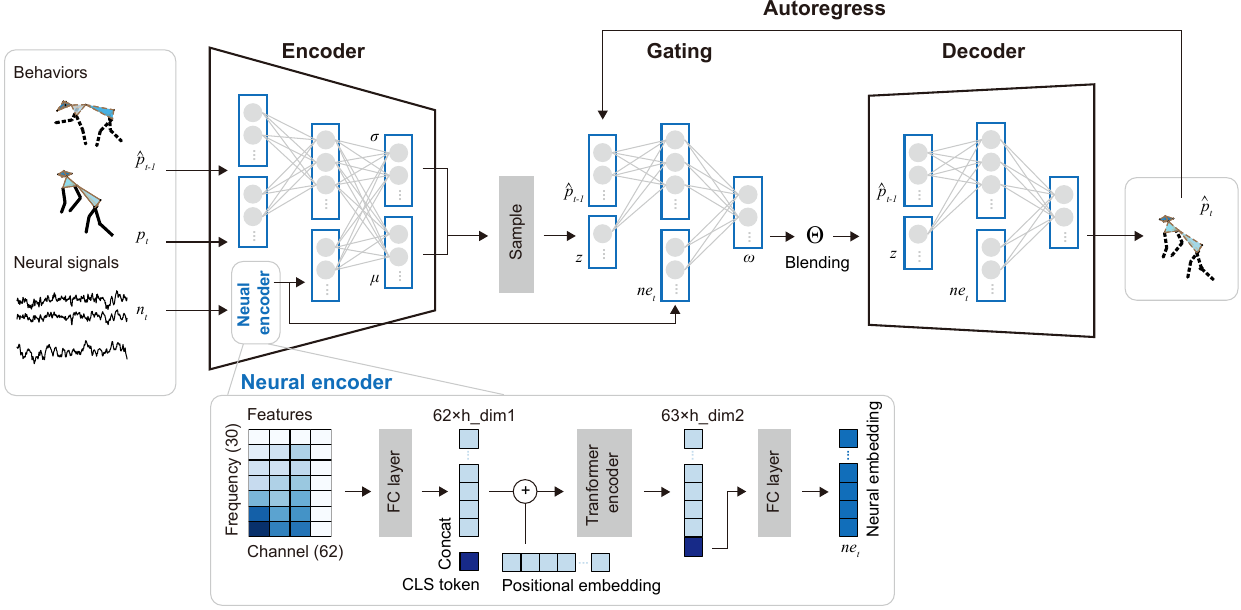}
    \caption{Framework of Neural-Behavioral Model. During training, the neural signal $n_t$ was first preprocessed and transformed into spectral features. These features were then encoded by the neural encoder, and together with the previously predicted movement state $\hat{p}_{t-1}$ (except for the initial state) and the current movement state $p_t$, were compressed by the encoder into a latent space $(\mu, \sigma)$. The decoder was implemented as a mixture-of-experts neural network. A latent variable $z$ was sampled from $(\mu, \sigma)$, and combined with the neural representation $ne_t$ and the previous movement state $\hat{p}_{t-1}$. These inputs were fed into the decoder to autoregressively reconstruct the current movement state $\hat{p}_t$. During inference on the test set, only the decoder was used. A latent variable $z$ was sampled from a standard Gaussian distribution, and together with the current neural representation $ne_t$ and the previously predicted movement state $\hat{p}_{t-1}$, the decoder autoregressively predicted the current movement state $\hat{p}_t$.}
    
  \label{Fig:network}
\end{figure}

\subsection{Neural-Behavioral model}

\paragraph{Latent variable dynamics model} Our goal is to model the probability distribution over state time series. We modeled the state transition process in which the transition model reconstructs the next state conditioned on the current behavior $p_{t-1}$, neural activity $n_{t}$  and a latent variable $z_{t}$ . For the behavior-only representation model (Behavioral Model) that does not incorporate neural features, the term $n_{t}$ is omitted. $\theta$ denotes the learnable parameters. 

\begin{equation}
  P_\theta(p_0,p_1,...,p_t) = P_\theta(p_0) \prod_{t=1}^{T} P_\theta(p_t \mid p_{t-1}, n_t)
  \label{define}
\end{equation}

\begin{equation}
  P_\theta(p_t \mid p_{t-1}, n_t)=\int P_\theta(z_t\mid p_{t-1},n_t)\,P_\theta(p_{t} \mid z_t,p_{t-1},n_t)\,dz_t
  \label{define}
\end{equation}
During training, the latent variable was regularized to follow a standard Gaussian distribution. The ground-truth state $p_t$ was used during optimization, and the encoder estimated the posterior distribution $q_\phi$. The reconstruction loss was defined as the mean squared error (MSE) between the predicted and the ground-truth movement representations. The network was trained using the standard evidence lower bound (ELBO) \cite{kingma2014auto} objective with the optimization target defined as follows.

\begin{equation}
    \mathcal{L}=\|p_{t}-\hat{p}_{t}\|^2+\beta  D_{KL} \left( q_\phi(z_t\mid p_{t-1},p_{t},n_{t})  \;\|\; \mathcal{N}(0,I) \right) 
  \label{define}
\end{equation}

At the inference stage, the model used the sampled $z_t$ from a standard Gaussian distribution, previously predicted $\hat{p}_{t-1}$ and the current neural activity $n_t$ to autoregressively generate movements step by step.

\paragraph{Neural-behavioral network} The architecture of neural-behavioral network is illustrated in Fig. \ref{Fig:network}. The encoder consists of a two-layer MLP for encoding behavior inputs and a neural encoder (See Appendix for details) for neural signals. The behavioral and neural features were concatenated and passed through a single MLP to project them into a latent space of dimension 16. The decoder comprised six expert networks with identical structures and a gating network. The gating network was implemented as a three-layer MLP. The use of a mixture-of-experts design helped to alleviate posterior collapse \cite{ling2020character}. Based on our empirical observations, neural signals were difficult to reconstruct accurately from a standard Gaussian latent space. Therefore, we did not include a neural reconstruction loss during training.

\paragraph{Data splits} We conducted computational experiments using data (18,000 samples, 30 minutes) collected during the first session of the first behavioral experiment day. The first 20 minutes of data were used as the training set, while the remaining 10 minutes were used as the test set.

\subsection{Reconstructing whole-body movement}

\begin{figure}
  \centering
  \includegraphics[width=\textwidth]{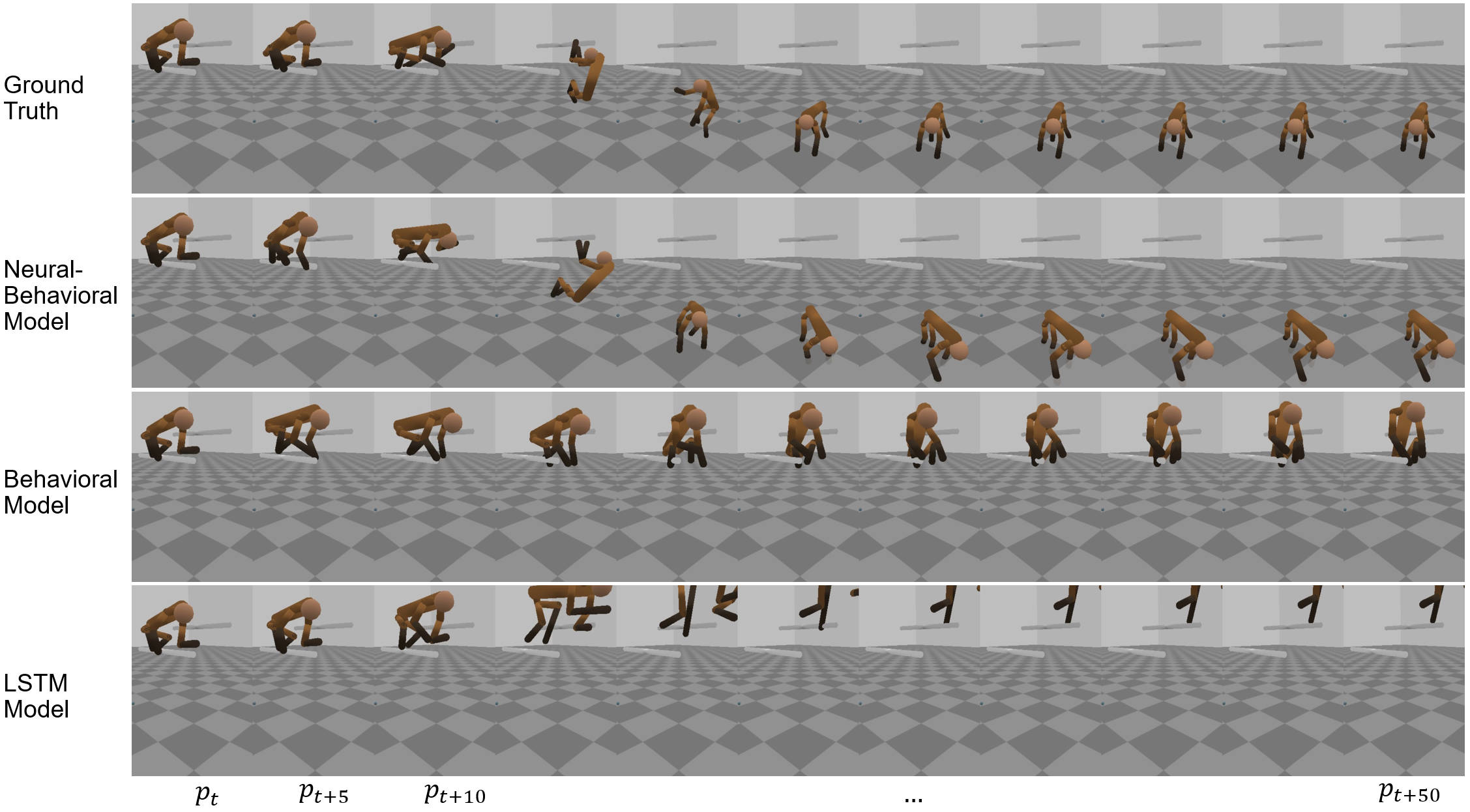}
  \caption{Visualization of reconstructed whole-body movement by three models during the monkey’s swinging movement from the upper bar. The $p_t$ denotes the given initial state. Models predict the movement at each time step, and the predicted movements at every 5 time steps (0.5 s) are visualized. Note that Neural-Behavioral Model yields movements closest to the ground truth.}
  \label{Fig:prediction result}
\end{figure}

To visualize model performance, Fig. \ref{Fig:prediction result} presents a representative example of joint kinematic predictions from different models during the monkey’s swinging movement from the upper bar to the floor (joint trajectories shown in Fig. \ref{Fig:traj} in the Appendix). We first evaluated the Neural-Behavioral Model, which reconstructed the current movement state from neural signals and the predicted state at the previous time step (except for the initial state). We observed that during the first 2 seconds (20 time steps), the model accurately reconstructed the monkey’s whole-body movement, despite slight jitter and minor temporal misalignment. Although after ~3 seconds, the accumulated error led to larger deviations in the reconstructed movement, the model still preserved a consistent trend of movement.

To demonstrate the necessity of neural signals in the movement reconstruction, we evaluated a Behavioral Model without neural signals. In this case, the model primarily generated random but plausible movement sequences that reflected general behavior priors, rather than the actual movement of the monkey.

We adopted a previously reported method \cite{eyvazpour2026general} that directly predicted movement from historical neural signals using an LSTM-based network and achieved state-of-the-art performance on constrained movement datasets \cite{chao2010long}. We noted that the LSTM model failed to predict motion direction in the world coordinate system. Therefore, all LSTM results were obtained by combining the predicted velocity magnitude with the direction of ground-truth velocity. In our dataset, the LSTM model failed to reconstruct accurate and physically consistent whole-body movement. Moreover, without behavior priors, it also failed to maintain physically consistent and stable postures.

The reconstruction performance of our Neural-Behavioral Model is largely independent of the initial state. When initialized from a stationary state, the reconstructed movement remains static (such as sitting, see Fig. \ref{Fig:other_result} in the Appendix). The model could also capture transitions such as initiating movement after a period of rest or returning to rest following movement (Fig. \ref{Fig:other_result}). However, it struggled to consistently reconstruct fine-grained movements, particularly in cases where most body parts remained static while only a subset (e.g., the wrist) was active.

\subsection{Performance evaluation}

We used the Average Displacement Error (ADE) and Final Displacement Error (FDE) to evaluate movement errors under different prediction time scale. The ADE is defined as the average MSE between reconstructed and ground-truth movement over all time steps, while FDE measures the MSE at the final time step. In addition, we computed the Pearson correlation coefficient between predicted and ground-truth joint trajectories to assess the global consistency of movement trends.

We manually selected 10 time points as initial states for movement reconstruction, rather than sampling randomly, since the monkey remained stationary for most of the test set. From each initial state, we autoregressively reconstructed 100 time steps (10 seconds). These 10 sequences covered several types of behaviors including sustained movement, transitions from movement to rest or conversely, and mostly static postures with slight limb actions. To ensure robustness, for each initial state we sampled the latent variable 5 times and reported the averaged performance.

\begin{table}[h]
  \caption{Performance of various models in predicting movements. Average estimation error of joints over 3 seconds. The unit of ADE and FDE was cm.}
  \label{Table: performance}
  \centering
  \begin{tabular}{lllllllll}
    \toprule
    Model & \multicolumn{3}{c}{Horizontal movement} & \multicolumn{3}{c}{Vertical movement} \\
    \cmidrule(r){2-4} \cmidrule(r){5-7} 
     & ADE & FDE  & r & ADE & FDE  & r &  \\
    \midrule
    Neural-Behavioral & \textbf{6.97} & \textbf{14.94} & \textbf{0.47} & \textbf{6.18} & \textbf{13.56}  & \textbf{0.40} \\
    Behavioral  & 10.54 & 23.62 & 0.11 & 10.56 & 28.54 & 0.20\\
    LSTM\cite{eyvazpour2026general}  & 16.93 & 32.61 & 0.11 & 12.39 & 34.90 & 0.02 \\
    \bottomrule
  \end{tabular}
  
\end{table}

 In Table \ref{Table: performance}, we compared prediction results between the Neural-Behavioral Model and the Behavioral Model. Compared with previous studies on human datasets that were mostly dominated by horizontal movements, we analyzed performance separately for horizontal and vertical movements, since monkeys exhibited rich vertical behaviors such as climbing and swinging. We also compared the movement prediction by the state-of-art LSTM model that was used for decoding under constrained settings. Notably, this is the first study to predict natural whole-body movement of freely moving monkey using population neuronal recordings, the performance of our Neural-Behavioral Model serves as a baseline for future research in this area.

In a previous work \cite{borgognon2025regional} that examined trained monkeys performing regular walking on a treadmill (rather than natural behaviors), neural signals recorded by microelectrode arrays were used to decode velocities of unilateral hindlimb joints during locomotion. This study reported a Pearson correlation coefficient of approximately 0.5 for predicting velocities of 4 joints. In contrast, our method achieved an average performance of whole-body joint positions with Pearson correlation coefficients 0.45 across all natural movements. Furthermore, we found that the Neural-Behavioral Model consistently outperformed the Behavioral Model, and both approaches significantly outperformed the LSTM model.

\begin{figure}
  \centering
  \includegraphics[width=\textwidth]{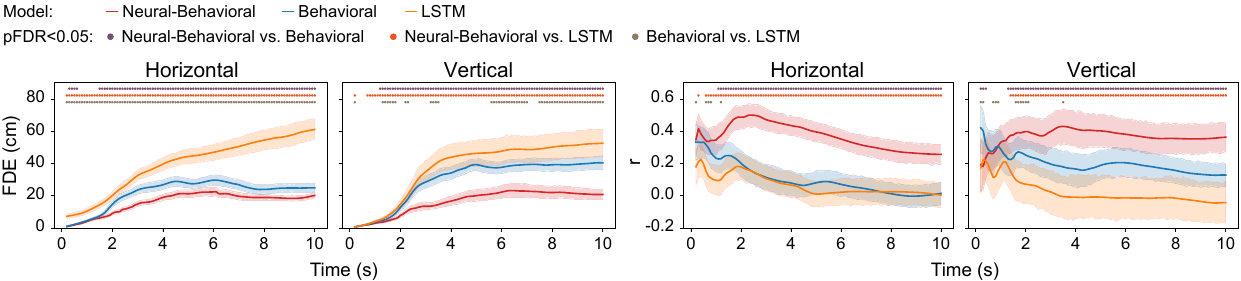}
  \caption{Model performance across various prediction time scales. Curves show the correlation coefficient FDE (left) and r (right) along the horizontal and vertical movement axes as functions of number of prediction steps. Shaded regions indicate 95\% confidence intervals across 15 joints and 10 movements, with colors denoting data from three different models. Corrected P-values < 0.05 are marked with colored dots above each plot for pairwise model comparisons.}
  \label{Fig:line plot}
\end{figure}

The Fig. \ref{Fig:line plot} provides a detailed comparison of the predictions by the three models for various joints and movements over 1–100 prediction steps (corresponding to 0–10 s). Across both horizontal and vertical movement axes, the Neural-Behavioral Model achieved overall better decoding performance compared with the Behavioral and LSTM models, as reflected by higher correlation coefficients (r) and lower FDE (see Fig. \ref{Fig:fde} in the Appendix for prediction FDE across various joints and movements along 3 axes). Statistical significance between model pairs at each number of steps was assessed using paired Wilcoxon signed-rank tests across matched joint-movement conditions, followed by false discovery rate (FDR) correction for multiple comparisons.

In addition, the relative advantage of the models depended on the temporal scale of prediction. At smaller numbers of prediction steps (within ${\sim}$1 s), the Neural-Behavioral Model showed no significant differences from the other two models in some situations (e.g., FDE and r for vertical movements), and even performed worse than the Behavioral Model in terms of r during vertical movements within 400 ms. However, as the number of prediction steps increased beyond about 1.5 s, performance of Neural-Behavioral Model always exceeded that of the other two models, showing as less FDE and higher r. These results indicate that Neural-Behavioral Model effectively integrates neural representations with behavior priors and learns a meaningful mapping between neural signals and behaviors. Whereas the Behavioral Model could achieve performance comparable to that of the Neural-Behavioral Model in short-term predictions (${\sim}$1 s), it failed to predict accurate movements over longer time scales. In contrast, the LSTM-based neural decoder developed under constrained settings failed to generalize to natural whole-body movements. Although it could capture coarse movement trends (see Fig. \ref{Fig:lstm} in the Appendix for predicted sacrum joint movement), it could reconstruct neither accurate nor physically plausible whole-body movements. We attribute the latter to the incapability of the LSTM-based model to deal with high-dimensional space of whole-body movements and the increased noise present in neural signals under natural conditions.

\section{Conclusion}

In this work, we addressed the problem of reconstructing natural whole-body movement from cortical activity in freely moving primates, which is substantially more challenging than conventional motor decoding. 
We developed an experimental platform that synchronously records large-scale epidural cortical activities and multi-view 3D whole-body kinematics of freely moving monkeys. We further formulated natural whole-body decoding as a neural-conditioned autoregressive generative modeling problem. The proposed Neural-Behavioral Model integrates cortical representations with a learned behavior prior, allowing the decoder to reconstruct continuous, coherent, and realistic whole-body movements directly from neural signals without imposing explicit physical constraints. It achieved superior performance compared with approaches that are based solely on neural representations or behavior priors. This work provides a proof-of-concept demonstration that natural whole-body movement in freely moving primates can be reconstructed from large-scale cortical activity with behavior priors. We release the original monkey neural-behavioral data to support future research on neural representation learning and motor brain-machine interfaces. Future work with larger longitudinal datasets, multiple animals, and more expressive neural-behavioral models may further improve long-horizon prediction and generalization.


\newpage
\renewcommand*{\bibfont}{\small}
\bibliographystyle{unsrt}
\bibliography{References}

\newpage


\appendix

\section{Appendix}

\renewcommand{\thefigure}{A\arabic{figure}}
\setcounter{figure}{0}

\renewcommand{\thetable}{A\arabic{table}}
\setcounter{table}{0}

\subsection{Representative time–frequency features of neural signals}
\begin{figure}[h]
  \centering
  \includegraphics[width=\textwidth]{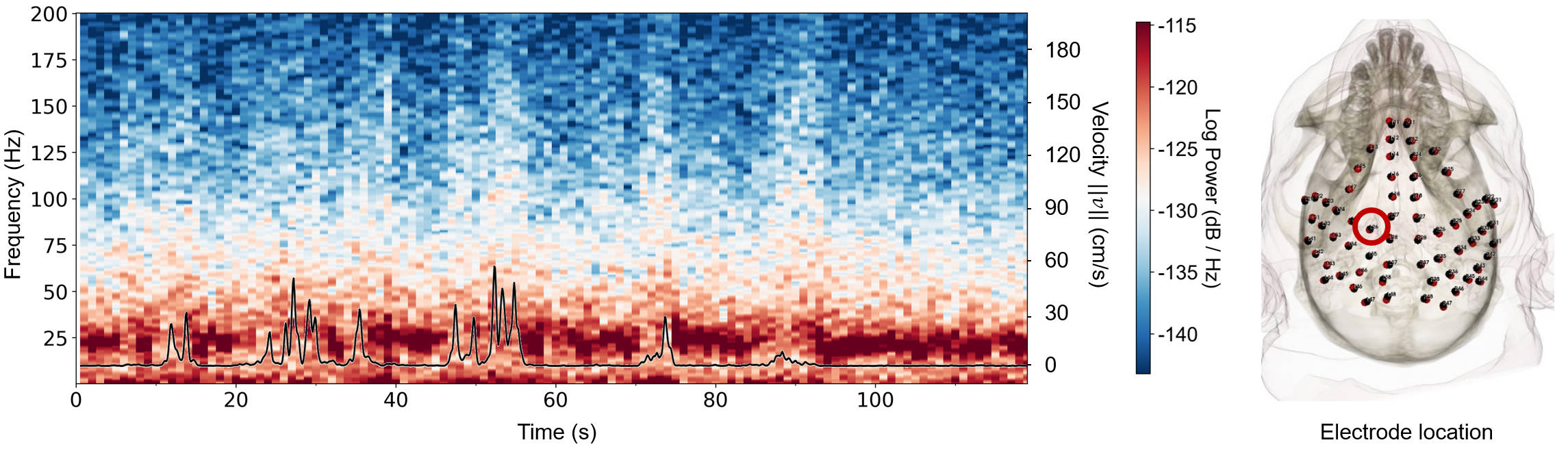}
    \caption{Time–frequency features of neural signals from the primary motor cortex. The time–frequency features are plotted together with the velocity magnitude of the right ankle in the world coordinate system (left panel). The right panel shows the electrode implantation locations (left primary motor cortex, red circles).}
  \label{Fig:PSD}
\end{figure}

We analyzed the time–frequency features of the primary motor cortex during the first 2 minutes of data (Fig. \ref{Fig:PSD}). By comparing the time–frequency features with the velocity magnitude of the right ankle, we observed an overall correlation between neural activity and movement. Specifically, during movement periods, beta-band de-synchronization and synchronization in the gamma and high-gamma bands were evident. However, compared to constrained movement condition \cite{farrokhi2020state}, this coupling was less consistent. For example, around 7 s and 22 s, no clear synchronous activity was observed. This suggests that neural activity during free behavior contains abundant spontaneous activity that is not directly related to movement. In addition, during 50–55 s, the time–frequency features failed to capture rapid transitions in movement, this may be one of the reasons why algorithms developed under constrained conditions failed to reconstruct natural whole-body movements. 

\subsection{Evaluation of monkey kinematic representation}

\begin{wrapfigure}{r}{0.5\textwidth}
    \vspace{-15pt}
    \centering
    \includegraphics[width=0.5\textwidth]{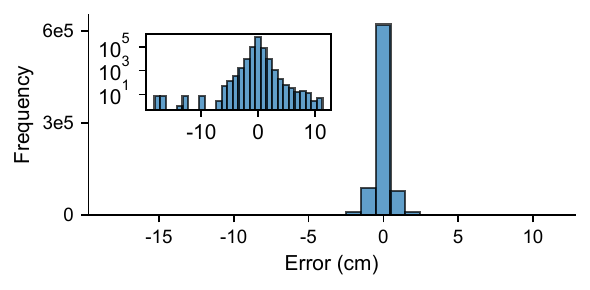}
    \vspace{-15pt} 
    \caption{Distribution of reconstruction errors for 3D coordinates from the kinematic representation. The y-axis denotes the total number of frames within each error bin. The upper-left inset is shown with a log-scaled y-axis.}
    \vspace{-5pt}
\label{Fig:error}
\end{wrapfigure}

\begin{figure}[h]
  \centering
  \includegraphics[width=\textwidth]{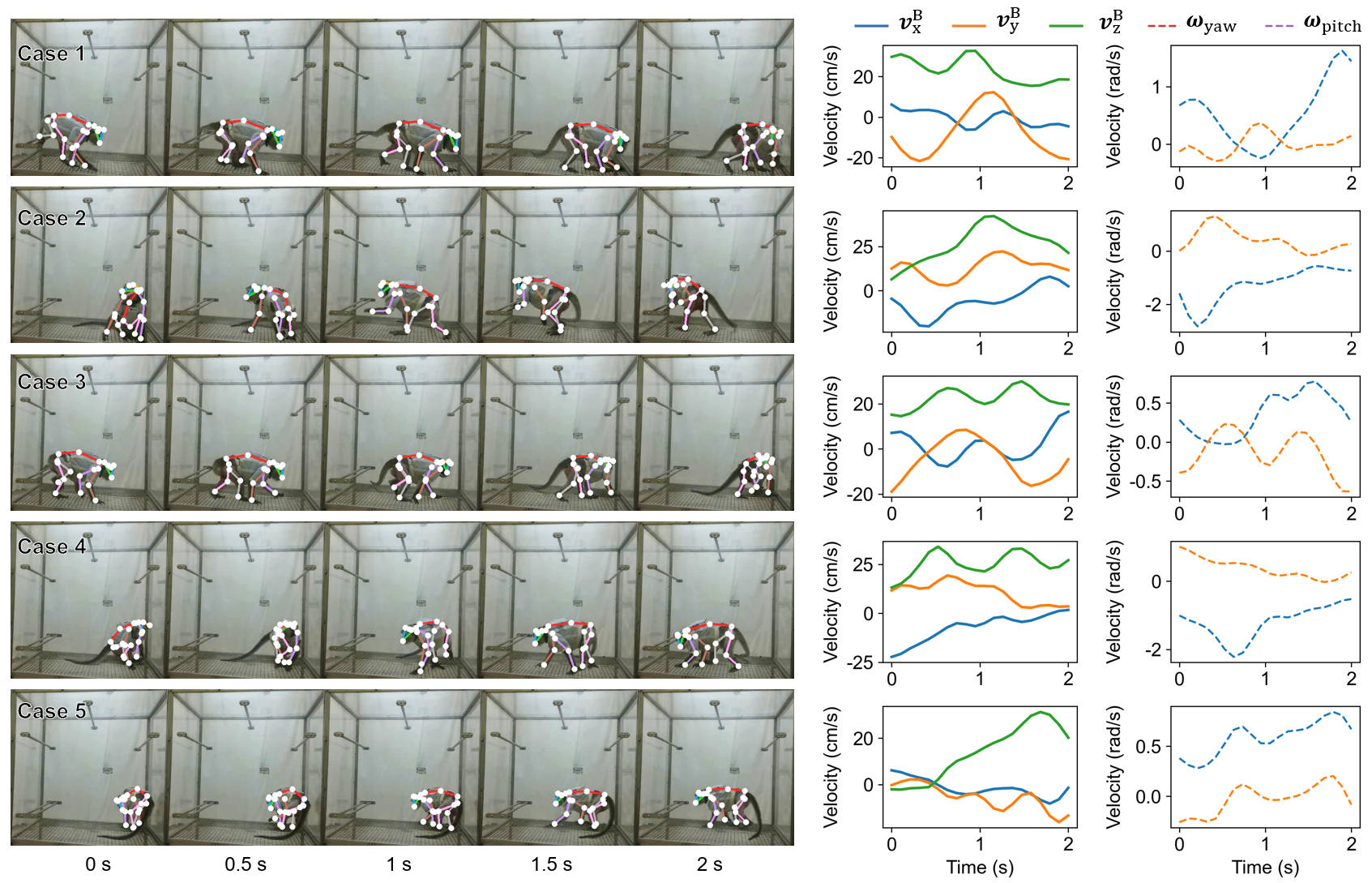}
    \caption{Example of monkey kinematic representation. Left: video frames from five walking segments with the 3D skeleton overlaid on the video. Case 1: 00:27–00:29, Case 2: 00:47–00:49, Case 3: 03:25–03:27, Case 4: 06:54–06:56, Case 5: 12:31–12:33. Right: the corresponding velocity of the root joint (sacrum), expressed in the root coordinate frame at the previous time step, for each walking case. yaw: rotation about the defined x-axis; pitch: rotation about the defined y-axis.}
  \label{Fig:kinematic_representation}
\end{figure}

In our kinematic representation, the reduction of joint degrees of freedom may introduce reconstruction errors. To evaluate the fidelity of the representation, we converted the 3D cartesian joint coordinates into the kinematic representation and then reconstructed the cartesian coordinates from this representation. We found that across all joints and all frames (10 Hz, totaling 18,000 frames), the mean reconstruction error was -0.012 cm, the mean absolute error was 0.30 cm, and the maximum error was 18.23 cm. The distribution of reconstruction errors is shown in Fig. \ref{Fig:error}, most error values are within 1 cm, indicating that the error remains within an acceptable range.

We computed the root-frame (sacrum joint) translation and rotation at each time step across 5 walking cases to characterize the monkey’s overall movement velocity, as shown in Fig. \ref{Fig:kinematic_representation}. Regardless of the walking direction, the translational velocity along the z-axis remained positive (indicating relative forward locomotion) and had a larger magnitude than the velocities along the x- and y-axes, suggesting that the z-axis represented the dominant movement direction. In contrast, the x- and y-axis velocities contained both positive and negative values, reflecting the variability of movement trajectories. For angular velocity, taking yaw rotation as an example, comparison between case 1 and case 2 showed that the monkey moved toward the left-front and right-front directions, respectively. The yaw velocity curves exhibited opposite rotational directions while maintaining similar temporal trends in velocity magnitude. Comparing case 1 and case 3, the overall movement direction was similar, but differences in angular velocity remained, reflecting variability in posture dynamics during movement. These results suggested that our kinematic representation captured meaningful movement characteristics of the monkey and facilitated the learning of appropriate behavior priors.

\subsection{Neural encoder detail}

\paragraph{Neural signal preprocessing and feature extraction} 
The signals from the 62 channels were originally sampled at 2000Hz. The blink-related electrooculographic components, separated using ICA, were verified for time alignment with the eyelid closure frames in the video. After confirming the time alignment, we applied filtering (1-300Hz), manual artifact removal, ICA-based artifact rejection, and re-referencing (averaging the left and right hemispheres separately). We extracted frequency-domain features as input to the neural encoder. Specifically, for each channel, the raw signal was first downsampled to 400 Hz. We then applied a sliding window with a step size of 100 ms (aligned with the movement data) and computed features over the past 0–500 ms segment, including multi-taper power spectral density (PSD) and complex Morlet wavelet transforms. We divided the 1–150 Hz frequency range into 15 log-space bins and averaged the wavelet coefficients within each bin, and divided the same frequency range into 15 linear-space bins and computed the corresponding averaged multi-taper PSD. The resulting representation formed a feature matrix of size $\mathbb{R}^{C \times 30}$, where $C$ denoted the number of recording channels.

\paragraph{Transformer-based neural encoder} To extract movement-related representations from noisy neural signals, we employed a Transformer-based neural encoder that mapped multi-channel neural activity into a compact latent representation as shown in  Fig. \ref{Fig:network}, and captured both spatial dependencies across electrode channels and spectral patterns in the frequency domain. For each time window, the neural input was represented as $N \in \mathbb{R}^{C \times F} $, where $C$ denoted the number of recording channels and $F$ denoted the number of frequency-band features extracted per channel. Each row corresponded to the energy or statistical features of a single channel across multiple frequency bands. The network first applied a linear projection layer to map the raw spectral features into a hidden dimension. Since each channel corresponded to a fixed electrode location in physical space, we further added learnable positional embeddings along the channel dimension to explicitly encode the spatial layout of the electrodes. A learnable global token was then added to the sequence to aggregate information across all channels. The resulting sequence was processed by standard Transformer blocks to model inter-channel dependencies. Finally, a regression head (LayerNorm followed by an MLP) maped the global representation to a movement-related latent variable.


\subsection{Training strategy}
To improve the stability of long-term movement generation in the autoregressive model, we divided the training process into three stages. During the initial stage (teacher), the model exclusively used the ground-truth $p_{t-1}$ and $p_t$ as inputs. This stage enabled the model to quickly learn stable temporal representations and accelerate convergence. During the intermediate transition stage (ramping), the input $p_{t-1}$ was updated using either the ground-truth movement or the model-predicted movement, determined randomly with a probability that gradually increased from 0 to 1. In the final stage (student), the model recursively generated motions using the predicted $p_{t-1}$ together with the ground-truth $p_t$, corresponding to fully autoregressive training. In the final training configuration, the three stages were trained for 30, 30, and 100 epochs, respectively.

\begin{figure}[h]
  \centering
  \includegraphics[width=\textwidth]{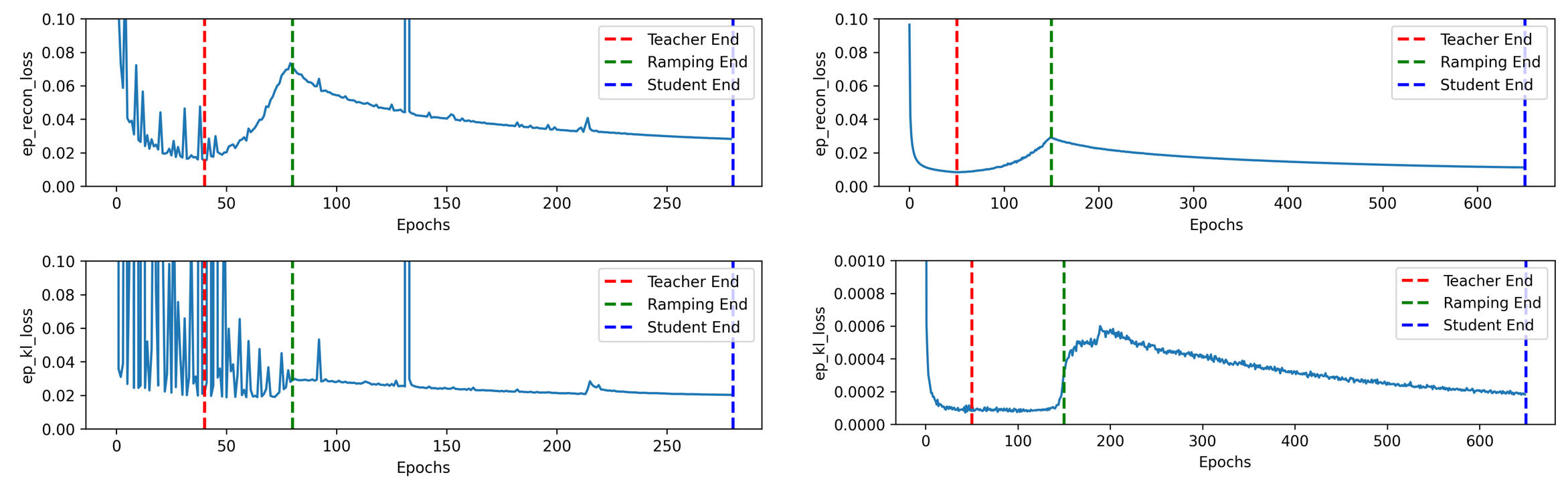}
  \caption{Example of train process. Loss curves before (left) and after (right) applying gradient-based constraints. We find that after applying the constraints, the training process remains stable over longer epochs, and the KL loss converges more rapidly.}
  \label{Fig:train process}
\end{figure}


We incorporated gradient-based constraints to regularize the training process and improve optimization stability. As shown in the Fig. \ref{Fig:train process}, applying the regularization significantly alleviated gradient-related issues and accelerated convergence during training. We used the reconstruction error on the test set to determine the number of training epochs, and the final model was trained for 160 epochs (taking approximately 1 hour on an NVIDIA GeForce RTX 4090 D 24GB GPU). 

\subsection{Comparison with related setup}

\begin{table}[h]
  \caption{Comparison of our data with related datasets. In the "Species" column, M denotes monkey, H human, and Mul. multiple. In the "Body parts" column, FL denotes forelimbs and HL denotes hindlimbs. The column "Label type" differentiates between 2D coordinates, 3D coordinates, and multiple (like EMG, behavior categories, and 3D shape representations). In the "Neural signal" column, MEA denotes micro-electrode array recording.}
  \label{Table: dataset}
  \centering
  \begin{tabular}{lllllll}
    \toprule
    Dataset  & Species   & Condition     &\multicolumn{2}{c}{Movement}  & Neural signal & Open \\
    \cmidrule(r){4-5}
      &   &   & Body parts & Label type  &  \\
    \midrule
    Ours   & M  & Natural & 21 joints  & 3D  & Epidural & Yes          \\
    UniMo4D\cite{wang2026x}   & Mul.& Natural & 25 joints  & 3D  & No & Part\\ 
    BigMaQ\cite{martini2026bigmaq} & M & Natural & 20 joints & Multiple & No & Yes \\ 
    O.M.S\cite{bala2020automated}   & M  & Natural & 13 joints  & 3D  & MEA & Yes          \\
    G.C.\cite{borgognon2025regional}  & M  & Constrained & 10 joints (HL) & Multiple  & MEA & No \\    
    NT\cite{bala2020automated} & M  & Constrained & 6 joints (FL)  & 2D  & Epidural & Yes \\
    
    \bottomrule
  \end{tabular}
  
\end{table}


\subsection{Additional results}

\begin{figure}[h]
  \centering
  \includegraphics[width=\textwidth]{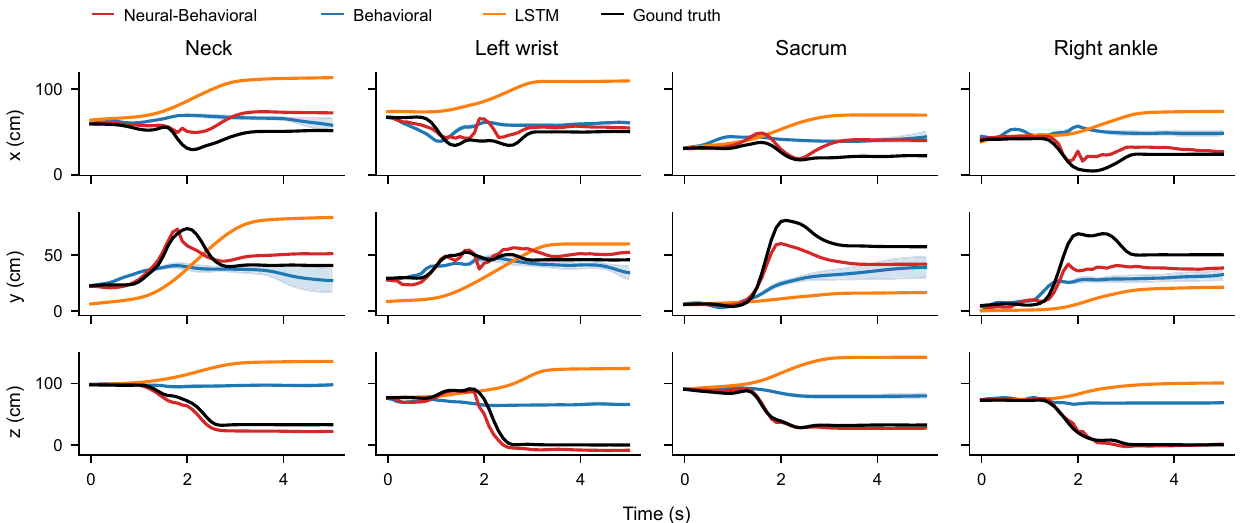}
  \caption{Trajectory comparison between ground truth (GT) and predictions from the three models. For the x, y, and z coordinates during a single swing-down movement (consistent with Fig \ref{Fig:prediction result}), across four joints. The Neural-Behavioral Model more closely tracks the GT trajectories than the other two models, particularly along the y and z axes, where kinematic variations are more pronounced. The LSTM model shows the largest deviation from GT across all joints and axes. Line colors indicate different models, and shaded regions denote the standard deviation 5 repeated runs.}
  \label{Fig:traj}
\end{figure}

\begin{figure}[h]
  \centering
  \includegraphics[width=\textwidth]{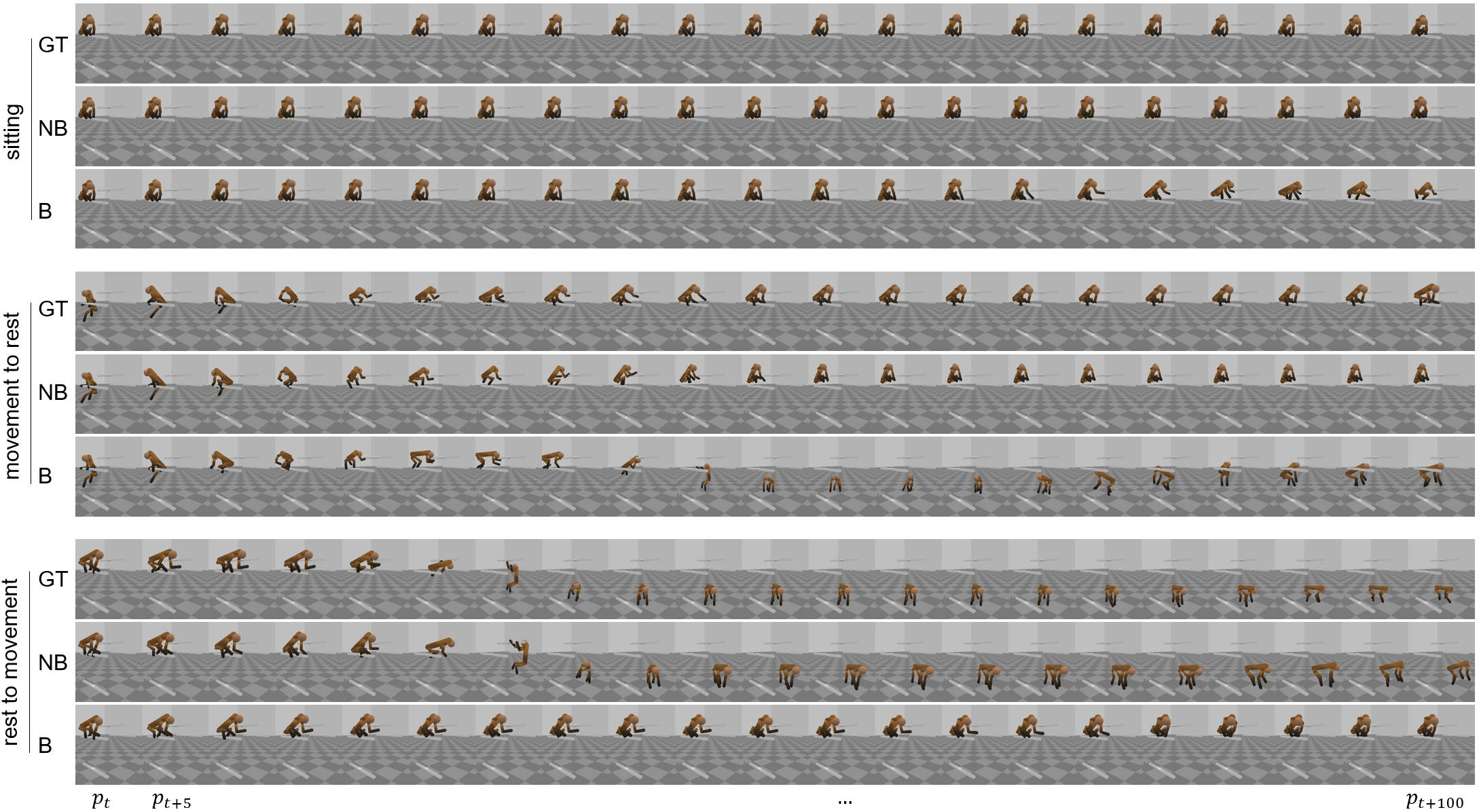}
  \caption{Visualization of additional reconstructed whole-body movements. $p_t$ denotes the given initial state, and the predicted movements at every 5 time steps (0.5 s) are visualized. GT: ground truth, NB: Neural-Behavioral Model, B: Behavioral Mode.}
  \label{Fig:other_result}
\end{figure}

\begin{figure}[h]
  \centering
  \includegraphics[width=\textwidth]{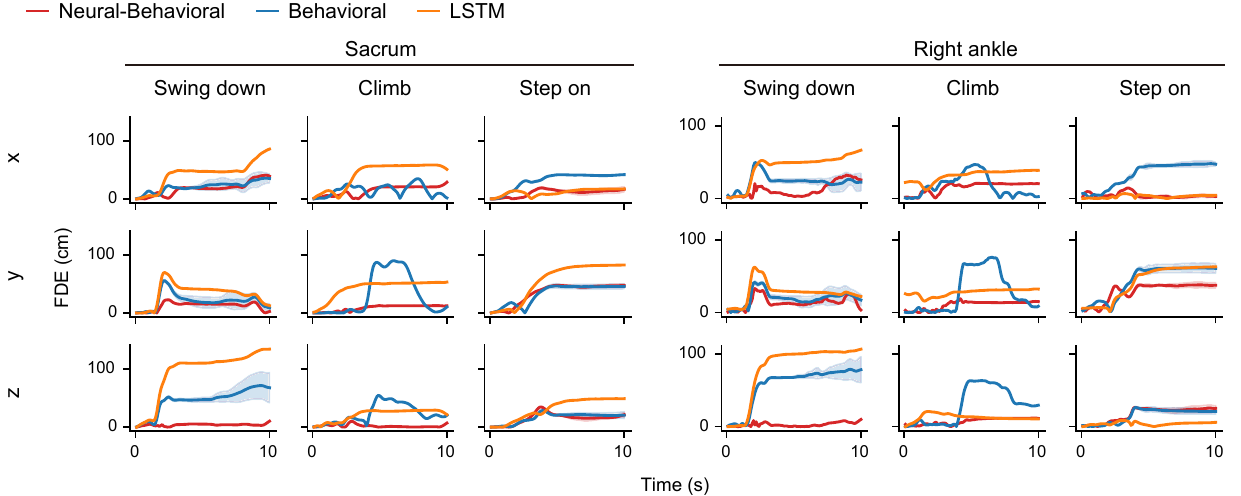}
  \caption{Prediction FDE of three models. Shown are x, y, and z axes of the “Swing down” “Climb” and “Step on” movements for the sacrum and right ankle joints. Overall, the Neural-Behavioral Model exhibits lower errors than the other two models. The improvement is especially obvious along the z-axis during “Swing down” and "Climb", along the y-axis during "Climb", as well as along the y-axis during “Step on” for the right ankle corresponding to the dominant movement directions. Line colors indicate different models, and shaded regions represent the standard deviation over 5 repeated runs. “Swing down”: swinging down from the upper bar to the floor; "Climb": climbing to the upper bar; "Step on": approaching and stepping onto the lower bar.}
  \label{Fig:fde}
\end{figure}

\begin{figure}[h]
  \centering
  \includegraphics[width=\textwidth]{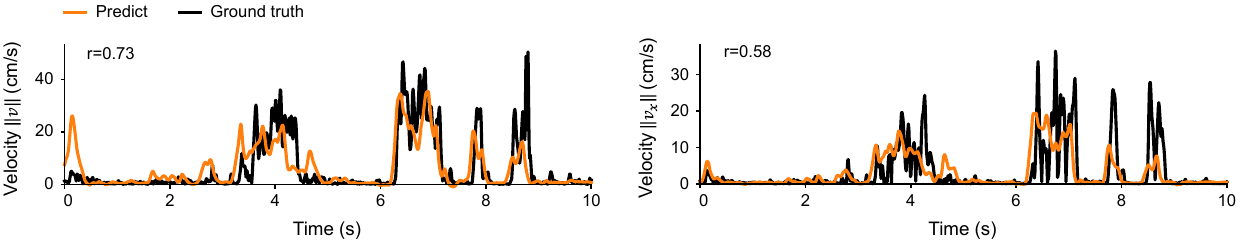}
  \caption{LSTM prediction. Using LSTM model to predict the 3D velocity vector magnitude, as well as velocity magnitude along the x-axis of the Sacrum joint in the monkey. The Pearson correlation coefficient r is shown in the upper-left corner of the figure. Note that all predicted velocities represent velocity magnitudes without directional information.}
  \label{Fig:lstm}
\end{figure}




\end{document}